\documentclass[letterpaper]{article}
\usepackage{times}
\usepackage[stable]{footmisc}
\usepackage{graphicx}
\usepackage{breakcites}
\usepackage{algorithm,algorithmic,a4wide,amssymb,multicol,enumitem,url}
\usepackage{amsmath}
\usepackage[breaklinks]{hyperref}
 \usepackage[top=3.7cm, bottom=3.7cm, left=3.7cm, right=3.7cm]{geometry} 

\def\xxx{\rotatebox[origin=c]{180}{RL} }
\def\xxxn{\rotatebox[origin=c]{180}{RL}}

\makeatletter
\renewcommand\hyper@natlinkbreak[2]{#1}
\makeatother
\newcommand{\mytilde}{\raise.17ex\hbox{$\scriptstyle\mathtt{\sim}$}}
\begin{document}

\title{
Reinforcement Learning Upside Down:  \\
Don't Predict Rewards - Just Map Them to Actions \\
{\small NNAISENSE/IDSIA Technical Report}
}

\date{23 June 2020 (based on version v1 of 5 Dec 2019) \\ 
{\small Earlier drafts:  21 Dec, 31 Dec 2017, 20 Jan, 4 Feb, 9 Mar, 20 Apr, 16 Jul 2018}}

\author{J\"{u}rgen Schmidhuber~\\
The Swiss AI Lab, IDSIA, USI  \& SUPSI  \\
NNAISENSE, Lugano, Switzerland \\
}
\maketitle

\begin{abstract}


We transform reinforcement learning (RL) into a form of supervised learning (SL) by turning traditional RL on its head, calling this \xxx or Upside Down RL (UDRL). Standard RL predicts rewards, while \xxx  instead  uses rewards as task-defining inputs, together with representations of time horizons and other computable functions of historic and desired future data. \xxx  learns to interpret these input observations as commands, mapping them to actions (or action probabilities)
 through SL on past (possibly accidental) experience.  \xxx generalizes to achieve high rewards or other goals,
through input commands such as:  {\em get lots of reward within at most so much time!}
\xxx can also learn to improve its exploration strategy. 
A separate paper~\cite{srivastava2019} on first experiments with \xxx shows that even a pilot version of  \xxx can
 outperform traditional baseline algorithms on certain challenging RL problems.

We also conceptually simplify an approach~\cite{levine2017} 
for teaching
a robot to imitate humans.
First videotape humans imitating the robot's current behaviors,
then let the robot learn through SL to map the videos (as input commands) to these behaviors,
then let it generalize and imitate videos of humans executing previously unknown behavior.  
This  {\em Imitate-Imitator} concept may actually explain why biological evolution has resulted in parents who imitate the babbling of their babies.

\end{abstract}

{\em Note: This is a minor update of recent work~\cite{udrl2019}}.

\newpage

\tableofcontents

\newpage
\section{Basic Ideas}
\label{basic}

Traditional RL machines~\cite{Kaelbling:96,Sutton:98,wiering2012} learn to predict rewards, given previous actions and observations, and learn to transform those predictions into rewarding actions. Our new method UDRL or  \xxx is radically different. It does not predict rewards at all. Instead it takes rewards {\em as inputs}. More precisely,  the \xxx machine observes
commands in form of  {\em desired rewards and time horizons}, such as: {\em``get so much reward within so much time.''} Simply by interacting with the environment,  it learns through gradient descent to map self-generated commands of this type to corresponding action probabilities. From such self-acquired knowledge it can extrapolate to solve new problems such as: {\em``get even more reward within even less time.''} 
Remarkably, a simple \xxx pilot version already outperforms traditional RL methods on certain challenging problems~\cite{srivastava2019}.

Let us outline this new principle in more detail.
An \xxx agent may interact with its environment during a single lifelong trial. At a given  time,  the history of actions and 
vector-valued~\cite{Schmidhuber:90diffgenau,Schmidhuber:90sandiego} costs (e.g.,  time, energy, pain \& reward signals) and other observations up to this time contains all the agent can know about the present state of itself and the environment. Now it is looking ahead up to some future horizon, trying  to obtain a lot of  reward until then. 

For all past pairs of times  {\em (time1 $<$ time2)}  it can 
retrospectively~\cite{hindsight2017,rauber2017hindsight}
invent additional, consistent, vector-valued {\em command} inputs for itself, indicating tasks such as: achieve the already observed rewards/costs between  {\em time1} and  {\em time2}. Or: achieve more than half this reward, etc.

Now it may simply use gradient-based SL to train a differentiable general purpose computer C such as a recurrent neural network  (RNN) \cite{Werbos:88gasmarket,WilliamsZipser:92,RobinsonFallside:87tr}\cite{888} to map the time-varying sensory inputs, augmented by the special {\em command} inputs defining time horizons and desired cumulative rewards etc, to the already known corresponding action sequences. 

If the experience so far includes different but equally costly action sequences leading from some start to some goal,  then C will learn to approximate the conditional expected values (or probabilities, depending on the setup) of appropriate actions, given the commands and other inputs. 

The single life so far may yield an enormous amount of knowledge about how to solve all kinds of 
problems with limited resources such as  time / energy / other costs.  
Typically, however, we want C to solve user-given problems, 
in particular, to get lots of reward quickly, e.g., by avoiding hunger (negative reward) caused by near-empty batteries, through quickly reaching the charging station without painfully bumping against obstacles. 
This desire can be encoded in a user-defined  command of the type  {\em (small desirable pain, small desirable time)}, and  C will generalize and act based on what it has learned so far through SL about starts, goals, pain, and time. This will prolong C's lifelong experience; all new observations immediately become part of C's growing training set, to further improve C's behavior in continual~\cite{Ring:94}
online fashion. 

For didactic purposes, we’ll first introduce formally  the basics of \xxx  for  deterministic environments and Markovian interfaces between controller and environment (Sec.~\ref{det}), then proceed to more complex cases in a series of additional Sections. 

A separate paper~\cite{srivastava2019} describes
 the concrete \xxx implementations used in our first experiments with 
 \xxxn, and presents remarkable  experimental results.

\section{Notation}
\label{formal}

More formally, in what follows, let $m$, $n$, $o$, $p$,  $q$,  $u$ denote positive integer constants, and $h$,
$i$, $j$, $k$,  $t$, $\tau$ positive integer variables assuming ranges implicit
in the given contexts.  The $i$-th component of any real-valued vector,
$v$, is denoted by $v_i$. 

To become a general problem solver that is able to run arbitrary
problem-solving programs, the controller C of an artificial
agent must be a general-purpose
computer~\cite{Goedel:31,Church:36,Turing:36,Post:36}.
Artificial recurrent neural networks (RNNs) fit this bill, e.g.,~\cite{888}.
The life span of 
our C (which could be an RNN)  can be partitioned 
into trials $T_1, T_2, \ldots$ However, possibly there is only one single, lifelong trial.  In each trial, C tries to manipulate some initially unknown environment through a sequence of
actions to achieve certain goals.
Let us consider one particular trial and its discrete sequence of
time steps, $t = 1,2,\ldots, T$.

At time $t$, 
during generalization of C's knowledge so far in Step 3 of Algorithm A1 or B1,
C receives as an input the concatenation of the following vectors:
a sensory input vector $in(t)\in \mathbb{R}^m$
(e.g., parts of $in(t)$ may represent the pixel
intensities of an incoming video frame),
a current 
vector-valued~\cite{Schmidhuber:90sandiego,Schmidhuber:91nips}
cost or reward vector $r(t)\in \mathbb{R}^n$
(e.g., components of $r(t)$ may
reflect external positive rewards, or negative values produced by pain
sensors whenever they measure excessive temperature or pressure or low battery load, that is, hunger), 
the previous output action $out'(t-1)$ 
(defined as an initial default vector of zeros in case of  $t = 1$; see below),
and extra variable task-defining input vectors
$horizon(t) \in \mathbb{R}^p$ 
(a unique and unambiguous representation of the current look-ahead time),
$desire(t) \in \mathbb{R}^n$
(a unique representation of the desired cumulative reward to be achieved until the end of the current look-ahead time),
and $extra(t) \in \mathbb{R}^q$ to encode 
additional user-given goals
(as we have done since 1990, e.g.,~\cite{Schmidhuber:91icannsubgoals,SchmidhuberHuber:91,powerplay2011and13}).

At time $t$, C then computes 
an output vector $out(t) \in \mathbb{R}^o$ 
used to select  the final output action $out'(t)$.
Often (e.g., Sec.~\ref{probdet}) 
$out(t)$ is interpreted as a probability distribution over possible actions.
For example,  
$out'(t)$ may be a {\em one-hot} binary vector $ \in \mathbb{R}^o$ with exactly one non-zero component,
$out_i'(t)=1$ indicates action $a^i$ 
in a set of  discrete actions $\{ a^1,a^2, \ldots, a^o \}$,
and $out_i(t)$ the probability of $a^i$.
Alternatively, for even $o$, $out(t)$ may encode the mean and the variance of 
a multi-dimensional Gaussian distribution over real-valued actions~\cite{Williams:92},
from which a high-dimensional action 
$out'(t) \in \mathbb{R}^{o/2}$ is sampled accordingly, e.g., to control a multi-joint robot. 
The execution of $out'(t)$ may
influence the environment and thus future inputs and rewards to C.

Let $all(t)$ denote the concatenation of $out'(t-1),in(t),r(t)$.
Let $trace(t)$ denote the sequence $(all(1), all(2), \ldots, all(t))$.


\section{Deterministic Environments With Markovian Interfaces}
\label{det}

For didactic purposes, we start with the  case of deterministic environments,
where there is a Markovian interface~\cite{Schmidhuber:91nips}
 between agent and environment,
such that C's current input tells C all there is to know about the
current state of its world.
In that case, C does not have to be an RNN - a multilayer feedforward network (FNN)~\cite{ivakhnenko1965,888}
is sufficient to learn a policy that maps inputs, desired rewards and time horizons to 
probability distributions over actions.

The following Algorithms A1 and A2 run in parallel, 
occasionally exchanging information at certain synchronization points. 
They make 
C learn many cost-aware policies from a single behavioral trace,
 taking into account many different possible time horizons.
 Both A1 and A2 use local variables  reflecting the input/output notation of Sec.~\ref{formal}.
 Where ambiguous, we distinguish local variables by appending the suffixes ``$[A1]$'' or ``$[A2]$,'' 
 e.g., $C[A1]$ or $t[A2]$ or $in(t)[A1]$.

\newpage

\subsection*{Algorithm A1: Generalizing through a copy of C (with occasional exploration)}
\label{alg}

\begin{enumerate}
\item  
Set $t:=1$. Initialize  local variable C (or $C[A1]$) of the type used to store controllers.

\item  Occasionally sync with Step 3 of Algorithm A2 to set  $C[A1]:=C[A2]$ 
(since $C[A2]$ is continually modified by Algorithm A2).

\item {\bf Execute one step:}  
Encode in $horizon(t)$ the goal-specific remaining time, e.g., until the end of the current trial
(or twice the lifetime so far~\cite{Hutter:05book+}).
Encode in $desire(t)$ a desired cumulative reward to be achieved within that time
(e.g., a known upper bound of the maximum possible cumulative reward,
or the maximum of (a) a positive constant and  (b) twice the maximum cumulative reward ever achieved before).
C observes the concatentation of  $all(t), horizon(t), desire(t)$ 
(and $extra(t)$, which may specify additional commands - see Sec.~\ref{lot} and Sec.~\ref{predicates}).
Then C outputs a 
probability distribution $out(t)$ over the next possible actions. 
Probabilistically 
select  $out'(t)$  accordingly
(or set it deterministically to one of the most probable actions).
In exploration mode (e.g., in a constant fraction of all time steps),
modify $out'(t)$ randomly (optionally, select  $out'(t)$ through some other scheme,
e.g., a traditional algorithm for planning or RL or black box optimization~\cite[Sec.~6]{888} - such details are not essential for \xxxn).
Execute action $out'(t)$
  in the environment, to get $in(t+1)$ and  $r(t+1)$.

\item Occasionally sync with Step 1 of Algorithm A2 to 
transfer the latest  acquired information about  $t[A1], trace(t+1)[A1]$,
to increase C[A2]'s training set through the latest observations. 

\item 
If the current trial is over, exit.
Set $t:=t+1$.  Go to 2.
\label{run}
\end{enumerate}

\subsection*{Algorithm A2: Learning lots of time \& cumulative reward-related commands}
\label{algA2}

\begin{enumerate}
\item Occasionally sync with  A1 (Step 4) to set $t[A2]:=t[A1]$, $trace(t+1)[A2]:=trace(t+1)[A1]$.

\item {\bf Replay-based training on previous behaviors and commands compatible with observed time horizons and costs:} 
{\bf For} all  pairs $\{(k,j); 1 \leq k \leq j \leq t\}$: 
train C through gradient descent-based  backpropagation~\cite{Linnainmaa:1970,Kelley:1960,werbos1982sensitivity}\cite[Sec.~5.5]{888}
to emit action $out'(k)$ at time $k$  in response to inputs  
$all(k)$, $horizon(k)$, $desire(k)$, $extra(k)$,
where  
$horizon(k)$ encodes the remaining time $j-k$ until time $j$,
and  $desire(k)$ encodes the total costs and rewards
$\sum_{\tau=k+1}^{j+1}r(\tau)$
 incurred through what happened between time steps $k$ and $j$.
(Here $extra(k)$ may be a non-informative vector of zeros - alternatives are 
discussed in Sec.~\ref{lot} and Sec.~\ref{predicates}.)
 
 \item  Occasionally sync with Step 2 of Algorithm A1 to 
copy  $C[A1]:=C[A2]$. Go to 1.

\label{replay}
\end{enumerate}

\subsection{Properties and Variants of Algorithms A1 and A2}
\label{commentsA}

\subsubsection{Learning Probabilistic Policies Even in Deterministic Environments}
\label{probdet}

 In Step 2 of Algorithm A2,
the past experience may contain 
many different, equally costly sequences of going from a state uniquely defined by $in(k)$ 
to a state uniquely defined by $in(j+1)$. 
Let us first focus on discrete actions encoded as {\em one-hot} binary vectors 
with exactly one non-zero component (Sec.~\ref{formal}).
Although the environnment is deterministic, 
by minimizing 
mean squared error (MSE), 
C will learn  {\em conditional expected values} 
\[
out(k)=E(out' \mid all(k), horizon(k), desire(k), extra(k)) 
\]
of corresponding actions, given C's inputs and training set,
where $E$ denotes the expectation operator.
That is, due to the binary nature of the action representation, 
C will actually learn to estimate {\em  conditional probabilities} 
 \[
out_i(k)=P(out'=a_i  \mid all(k), horizon(k), desire(k), extra(k)) 
\]
of appropriate actions, given C's inputs and training set. For example, in a video game, two equally long paths may have led from location A to location B around some obstacle, one passing it to the left, one to the right, and C may learn a 50\% probability of going left at a fork point, but afterwards there is only one fast way to B,  and  C can learn to henceforth move forward with highly confident actions, assuming the present goal is to minimize time and energy consumption.

\xxx is of particular interest for high-dimensional  actions
(e.g., for complex multi-joint robots), 
because SL can easily deal with those, while traditional RL does not. 
See Sec.~\ref{hidim} for learning probability distributions over such actions, 
possibly  with statistically dependent action components.

\subsubsection{Compressing More and More Skills into C}
\label{compressskills}
In Step 2 of Algorithm A2, more and more  skills are  compressed or collapsed into C, 
like in the chunker-automatizer system of the 1991 neural history compressor~\cite{chunker91and92}, 
where a student net (the ``automatizer'') is continually re-trained 
not only on its previous skills (to avoid forgetting), but also
 to imitate the behavior of a teacher net  (the ``chunker''), 
which itself keeps learning new behaviors.

\subsubsection{No Problems With Discount Factors}
\label{disount}

Some of the math of traditional RL~\cite{Kaelbling:96,Sutton:98,wiering2012} heavily relies on problematic discount factors. 
Instead of maximizing
$\sum_{\tau=1}^Tr(\tau)$, 
many RL machines try to maximize 
$\sum_{\tau=1}^T\gamma^\tau r(\tau)$
or
$\sum_{\tau=1}^{\infty}\gamma^\tau r(\tau)$ (assuming unbounded time horizons),
where the positive real-valued discount factor 
$\gamma < 1$ distorts the real rewards in exponentially shrinking fashion,
thus simplifying certain proofs (e.g., by exploiting that $\sum_{\tau=1}^{\infty}\gamma^\tau r(\tau)$ is finite).

\xxxn, however, explicitly takes into account observed time horizons in a precise and natural way, does not assume infinite horizons, and does not suffer from distortions of the basic RL problem.

\subsubsection{Representing Time / Omitting Representations of Time Horizons}
\label{time}
What is a good way of representing look-ahead time through 
$horizon(t) \in \mathbb{R}^p$?
The simplest way may be $p=1$ and $horizon(t) =t$. 
A less quickly diverging representation is $horizon(t) =\sum_{\tau=1}^{t}1/\tau$.
A bounded representation is $horizon(t) =\sum_{\tau=1}^{t}\gamma^{\tau} \tau$ with positive real-valued $\gamma<1$.
Many distributed representations with $p>1$ are possible as well, e.g., date-like representations. 


In cases where C's life can be segmented
into several time intervals or {\em episodes} of varying lengths unkown in advance,
and where we are only interested in C's total reward per episode, 
we may omit C's $horizon()$-input. C's  $desire()$-input still can be
 used to encode the desired cumulative reward until the time when 
  a special component of C's $extra()$-input  switches from 0 to 1, 
 thus indicating the end of the current episode. 
 It is straightforward to modify 
 Algorithms A1/A2 accordingly.

\subsubsection{Computational Complexity}
\label{complexity}
The replay~\cite{Lin:91} of Step 2 of Algorithm A2 can be done in $O(t(t+1)/2)$ time per training epoch.
In many real-world applications, 
such quadratic growth of computational cost  may 
be negligible compared to the costs of executing
actions in the real world.
(Note also that hardware is still getting exponentially cheaper over time, 
overcoming any simultaneous quadratic slowdown.) See Sec.~\ref{reducing}.

\subsubsection{Learning a Lot From a Single Trial - What About Many Trials?}
\label{lot}
In Step 2 of Algorithm A2, for every time step, C learns to obey many commands of the type:
get so much future reward within so much time. That is,
from a single trial of only 1000 time steps, it derives roughly half a million 
trainig examples conveying a lot of fine-grained knowledge about time and rewards.
For example, C may learn that small increments of time often correspond to small increments of costs and rewards, except at certain crucial moments in time, e.g., at the end of a board game
when the winner is determined. A single behavioral trace may thus inject an enormous amount of knowledge into C, which can learn to explicitly represent all kinds of long-term and short-term causal relationships between actions and consequences, given the initially unknown environment. For example, in typical physical environments, C could  automatically learn detailed maps of space / time / energy / other costs 
associated with moving from many locations (at different altitudes)  to many target locations~\cite{SchmidhuberHuber:91,Schmidhuber:91icannsubgoals,powerplay2011and13,hindsight2017,rauber2017hindsight} encoded as parts of $in(t)$ or of $extra(t)$ - compare Sec.~\ref{locations}.

If there is not only one single lifelong trial, we may 
run Step 2 of Algorithm A2  for previous trials as well, to avoid forgetting of previously learned skills,
like in the {\sc PowerPlay}  framework~\cite{powerplay2011and13,Srivastava2013first}.

\subsubsection{How Frequently Should One Synchronize Between Algorithms A1 and A2?}
\label{sync}
It depends a lot on the  task and the computational hardware. 
In a real world robot environment, 
executing a single action in Step 3 of A1 may take more time than billions of training iterations in Step 2 of A2. 
Then  it might be most efficient to  sync after every single real world action, 
which immediately may yield for C many new insights  into the workings of the world. 
On the other hand, when actions and trials are cheap, e.g., in simple simulated worlds,  it might be most efficient to
synchronize rarely.

\subsubsection{On Reducing Training Complexity by Selecting Few Relevant Training Sequences}
\label{reducing}


To reduce the complexity  $O(t(t+1)/2)$ of Step 2 of Algorithm A2  (Sec. \ref{complexity}), 
certain SL methods will ignore most of the training sequences defined by the pairs $(k,j)$ of Step 2,
and instead select only a few of them, either randomly, or by selecting {\em prototypical} sequences,
inspired by {\em support vector machines} (SVMs) whose only effective training examples
are the {\em support vectors} identified through a margin criterion~\cite{Vapnik:95,advkernel},
such that (for example) correctly classified outliers do not directly affect the final classifier. 
In environments where actions are cheap, the selection of only few training sequences may also allow for 
synchronizing more frequently between Algorithms A1 and A2 (Sec. \ref{sync}).

Similarly, 
when the overall goal is to learn a single rewarding behavior 
through a series of trials, 
at the start of a new trial, a variant of A2 could  simply 
delete/ignore   the training sequences collected during most of the less rewarding previous trials,
while Step 3 of A1 could still demand more reward than ever observed.
Assuming that C is getting better and better at acquiring reward over time,
this will not only reduce training efforts, but also bias C towards recent rewarding behaviors,
at the risk of making C forget how to obey commands demanding low rewards.  

There are numerous applicable SL {\em tricks of the trade} (e.g., ~\cite{tricksofthetrade:2012}) 
and sophisticated ways of selectively deleting past experiences from the training set
to improve and speed up  SL.

\section{Other  Properties of the History as Command Inputs}
\label{predicates}

A single trial can yield even much more additional information for C than what is 
exploited in Step 2 of Algorithm A2. For example, the following addendum to Step 2 trains C
to also react to an input command saying 
{\em ``obtain {\bf more than} this reward within so much time''}
instead of 
{\em ``obtain so much reward within so much time,''}
simply by training on all past experiences that retrospectively match this command.

\begin{enumerate}

\item[2b.] {\bf Additional replay-based training on previous behaviors and commands compatible with observed time horizons and costs for Step 2  of Algorithm A2:} 
{\bf For} all  pairs $\{(k,j); 1 \leq k \leq j \leq t\}$: 
train C through gradient descent
to emit action $out'(k)$ at time $k$  in response to inputs  
$all(k)$, $horizon(k)$, $desire(k)$, $extra(k)$,
where one of the components of $extra(k)$ is a
special  binary input $morethan(k):=1.0$
(normally 0.0),
where  
$horizon(k)$ encodes the remaining time $j-k$ until time $j$,
and  $desire(k)$ encodes {\em half}  the total costs and rewards
$\sum_{\tau=k+1}^{j+1}r(\tau)$
 incurred between time steps $k$ and $j$,
or 3/4 thereof, or 7/8 thereof, etc.
\label{morethan2b}
\end{enumerate}

That is, C now also learns to generate probability distributions over action trajectories that yield {\em more than} a certain amount of reward within a certain amount of time. Typically, their number greatly exceeds the number of trajectories yielding {\em exact} rewards, which will be reflected in the correspondingly reduced conditional probabilities of action sequences learned by C.  

A natural corresponding modification of Step 3 of Algorithm A1 is to 
encode in $desire(t)$ the maximum conditional reward ever achieved, given 
$all(t), horizon(t)$, and to
activate the special  binary input $morethan(t):=1.0$ as part of $extra(t)$,
such that C can generalize from what it has learned so far about the concept of obtaining
{\em more than} a certain amount of reward within a certain amount of time. 
{\bf Thus \xxx can learn to improve its exploration strategy in goal-directed fashion.}

\subsection{Desirable Goal States / Locations}
\label{locations}

Yet another modification of  Step 2 of Algorithm A2 is to 
encode within parts of $extra(k)$  a final desired input  $in(j+1)$ (assuming $q>m$),
like in previous work where extra inputs are used to define goals or  target locations~\cite{SchmidhuberHuber:91,Schmidhuber:91icannsubgoals,powerplay2011and13,hindsight2017,rauber2017hindsight},
such that C can be trained to execute commands of the type  
{\em ``obtain so much reward within so much time and finally reach a particular state identified by this particular input.''}
See Sec.~\ref{arbitrary} for generalizations of this.

The natural corresponding modification of Step 3 of Algorithm A1 is to encode such desired inputs~\cite{SchmidhuberHuber:91}  in $extra(t)$, e.g., a 
goal location that has never been reached before. 

\subsection{Infinite Number of Computable, History-Compatible Commands}
\label{infinite}

Obviously there are infinitely many other 
computable functions of subsequences of $trace(t)$ with binary outputs {\em true} or {\em false}  
that yield  {\em true}  when applied to certain subsequences.
In principle, such {\em computable predicates} could be encoded  in Algorithm A2 as unique commands for C with the help of $extra(k)$, 
to further increase C's knowledge about how the world works, 
such that C can better generalize when it comes to planning future actions in Algorithm A1. 
In practical applications, however, one can train C only on finitely many commands, which should be chosen wisely.  

Note the similarity to  {\sc PowerPlay}  
(2011)~\cite{powerplay2011and13,Srivastava2013first} which  allows for 
{\em arbitrary computable task specifications} as extra inputs to an RL system.
Since in general there are  many possible tasks, 
 {\sc PowerPlay}  has a built-in way of selecting new tasks automatically and economically.
 {\sc PowerPlay}, however, not only looks backwards in time to find new commands compatible with the observed history,
but can also actively set goals that require to obtain new data from the environment through interaction with it.

\section{Probabilistic Environments}
\label{probenv}

In probabilistic environments, for  two different time steps $l \neq h$
we may have  
$all(l)=all(h)$, $out(l)=out(h)$
but 
$r(l+1) > r(h+1)$,
due to {\em ``randomness''} in the environment. 
To address this, let us first discuss  {\em expected} rewards. 
Given $all(l), all(h)$ and
keeping the Markov assumption of Sec.~\ref{det},
we may use C's command input $desire(.)$ to encode
a desired  {\em expected} immediate reward of $1/2[r(l+1) + r(h+1)]$ 
which, together with $all(h)$ and a $horizon()$ representation of 0 time steps,  should be 
 mapped to  $out(h)$ by C,
 assuming  a uniform conditional reward distribution.

More generally, 
assume  a finite set of states $\{ s^1,s^2, \ldots, s^u \}$, 
each with an unambiguous encoding through C's $in()$ vector,
and actions $\{ a^1,a^2, \ldots, a^o \}$ with one-hot encodings (Sec.~\ref{formal}).
For each pair $(s^i,a^j)$ we can use a real-valued variable $z_{ij}$ to
estimate~\cite{hastie2009} the expected immediate reward for executing $a^j$ in $s^i$.
This reward is assumed to be independent of the history of previous actions and observations
(Markov assumption~\cite{stratonovich1960}).

$z_{ij}$ can be updated incrementally and cheaply whenever $a^j$ is executed in $s^i$  
 in Step 3 of Algorithm A1, and the resulting immediate reward is observed. 
The following simple modification of Step 2 of Algorithm A2 trains C
to map {\em desired expected} rewards (rather than plain rewards) to actions,
based on the observations so far.

\begin{enumerate}

\item[2*] {\bf Replay-based training on previous behaviors and commands compatible with observed time horizons and expected costs in probabilistic Markov environments for Step 2  of Algorithm A2:} 
{\bf For} all  pairs $\{(k,j); 1 \leq k \leq j \leq t\}$: 
train C through gradient descent
to emit action $out'(k)$ at time $k$  in response to inputs  
$all(k)$, $horizon(k)$, $desire(k)$  (we ignore $extra(k)$ for simplicity),
where  
$horizon(k)$ encodes the remaining time $j-k$ until time $j$,
and  $desire(k)$ encodes the {\em estimate of the  total  expected costs and rewards}
$\sum_{\tau=k+1}^{j+1}E(r(\tau))$, where the $E(r(\tau))$ are estimated in the obvious way through the $z_{..}$ variables
corresponding to visited states / executed actions between time steps $k$ and $j$.
\label{morethan2*}
\end{enumerate}

If randomness is affecting not only the immediate reward for executing $a^j$ in $s^i$
but also the resulting next state, then
{\em Dynamic Programming} (DP)~\cite{Bellman:1957} can still 
estimate in similar fashion cumulative expected rewards
 (to be used as command inputs encoded in $desire()$), given
the training set so far. This approach essentially adopts central aspects of  
traditional DP-based RL~\cite{Kaelbling:96,Sutton:98,wiering2012}
without affecting the method's overall order of computational complexity (Sec.~\ref{complexity}).

From an algorithmic point of view~\cite{Solomonoff:64,Kolmogorov:65,LiVitanyi:97,Schmidhuber:02ijfcs},
however,
randomness simply reflects a separate,
 unobservable oracle injecting extra bits of 
information into the observations.
Instead of learning to map {\em expected} rewards to actions as above,
C's problem of partial observability can also be addressed by adding to C's input  
a unique representation of the current time step, 
such that it can learn the {\em concrete}  reward's dependence on time,
and is not misled by a few lucky past experiences. 

It is most natural to consider  the case of probabilistic 
environments as a special case of partially observable environments 
discussed next  in Sec.~\ref{partial}.

\section{Partially Observable Environments}
\label{partial}

In case of a non-Markovian interface~\cite{Schmidhuber:91nips}
 between agent and environment,
C's current input does not 
tell C all there is to know about the
current state of its world.
A recurrent neural network  (RNN)~\cite{888} or a similar general purpose computer
may be required to translate the entire history of previous observations and actions
into a meaningful representation of the present world state.
Without loss of generality, we focus on C being an RNN such as 
LSTM~\cite{lstm97and95,Gers:2000nc,Graves:09tpami,888}
which has become highly commercial, 
e.g.,~\cite{googlevoice2015,wu2016google,amazon2016,facebook2017}.
Algorithms A1 and A2 above have to be modified accordingly, resulting in Algorithms B1 and B2
(with local variables and input/output notation analoguous to A1 and A2,  e.g., $C[B1]$ or $t[B2]$ or $in(t)[B1]$).  


\subsection*{Algorithm B1: Generalizing through a copy of C (with occasional exploration)}
\label{algB1}

\begin{enumerate}
\item  
Set $t:=1$. Initialize  local variable C (or $C[B1]$) of the type used to store controllers.

\item  Occasionally sync with Step 3 of Algorithm B2 to {\bf  do:} 
copy  $C[B1]:=C[B2]$ (since $C[B2]$ is continually modified by Algorithm B2). 
Run C on $trace(t-1)$, 
such that C's internal state contains
a memory of the history so far,
where the inputs $horizon(k)$, $desire(k)$, $extra(k)$, $1 \leq k < t$ are retrospectively 
adjusted to match the observed reality up to time $t$.
One simple way of doing this is to let $horizon(k)$ represent 0 time steps, 
$extra(k)$ the null vector, and to set $desire(k)=r(k+1)$, 
 for all $k$
(but many other consistent commands are possible, e.g., Sec.~\ref{predicates}).


\item {\bf Execute one step:}  
Encode in $horizon(t)$ the goal-specific remaining time (see Algorithm A1).
Encode in $desire(t)$ a possible future cumulative reward,
and in $extra(t)$ additional goals, e.g., to receive more than this reward within the
remaining time - see   Sec.~\ref{predicates}.
C observes the concatentation of  $all(t), horizon(t), desire(t), extra(t)$,
and  outputs  $out(t)$. 
Select  action $out'(t)$  accordingly.
In exploration mode (i.e., in a constant fraction of all time steps),
modify $out'(t)$ randomly. 
Execute $out'(t)$
  in the environment, to get $in(t+1)$ and  $r(t+1)$.

\item Occasionally sync with Step 1 of Algorithm B2 to 
transfer the latest  acquired information about  $t[B1], trace(t+1)[B1]$,
to increase C[B2]'s training set through the latest observations.

\item 
If the current trial is over, exit.
Set $t:=t+1$. 
Go to 2.
\label{runalgB1}
\end{enumerate}

\subsection*{Algorithm B2: Learning lots of time \& cumulative reward-related commands}
\label{algB2}

\begin{enumerate}

\item Occasionally sync with  B1 (Step 4) to set $t[B2]:=t[B1]$, $trace(t+1)[B2]:=trace(t+1)[B1]$.

\item {\bf Replay-based training on previous behaviors and commands compatible with observed time horizons and costs:} 
{\bf For} all  pairs $\{(k,j); 1 \leq k \leq j \leq t\}$ {\bf do}: If $k>1$,
run RNN C on $trace(k-1)$ to create an internal representation of the history up to time $k$,
where for $1 \leq i < k$, $horizon(i)$ encodes 0 time steps, $desire(i)=r(i+1)$,
and $extra(i)$ may be a vector of zeros (see  Sec.~\ref{predicates},~\ref{time},~\ref{arbitrary} for alternatives). 
Train RNN C 
to emit action $out'(k)$ at time $k$  in response to this previous history  (if any) and $all(k)$, 
where the special command input 
$horizon(k)$ encodes the remaining time $j-k$ until time $j$,
and  $desire(k)$ encodes the total costs and rewards
$\sum_{\tau=k+1}^{j+1}r(\tau)$
 incurred through what happened between time steps $k$ and $j$,
while $extra(k)$ may encode additional commands compatible with the observed history, e.g., 
Sec.~\ref{predicates},~\ref{arbitrary}.

 \item  Occasionally sync with Step 2 of Algorithm B1 to 
copy  $C[B1]:=C[B2]$. Go to 1.

\end{enumerate}

\subsection{Properties and Variants of Algorithms B1 and B2}
\label{commentsB}

Comments of Sec.~\ref{commentsA} apply in analaguous form, generalized to the RNN case.
In particular, 
although each  replay for some pair of time steps $(k,j)$  in Step 2 of Algorithm B2 takes into account the 
entire history up to $k$ and the subsequent future up to $j$,
Step 2 can be implemented such that its computational
complexity is still only $O(t^2)$ per training epoch
(compare Sec.~\ref{complexity}).

\subsubsection{Retrospectively Pretending a Perfect Life So Far}
\label{perfect}

Note that during generalization in Algorithm B1, 
RNN C always acts as if its life so far has been perfect, 
as if it always has achieved what it was told,
because its command inputs are retrospectively adjusted to match the observed outcome,
such that RNN C is fed with a consistent history of commands and other inputs.

\subsubsection{Arbitrarily Complex Commands for RNNs as General Computers}
\label{arbitrary}

Recall Sec.~\ref{predicates}. Since RNNs are general computers, 
we can train an RNN C on additional complex commands compatible with the observed history, 
using $extra(t)$ to help encoding commands such as: 
{\em ``obtain more than this reward within so much time, while visiting a particular state 
(defined through an extra goal input encoded 
in $extra(t)$~\cite{SchmidhuberHuber:91,Schmidhuber:91icannsubgoals}) at least 3 times, but not more than 5 times.''}

That is, like in  {\sc PowerPlay}  (2011)~\cite{powerplay2011and13}, we can train C to obey 
essentially arbitrary computable 
task specifications that match previously observed traces of actions and inputs. 
Compare Sec.~\ref{predicates}, \ref{infinite}.
(To deal with (possibly infinitely) many tasks,  {\sc PowerPlay} can order tasks 
by the computational effort required to add their solutions to the task repertoire.)

\subsubsection{High-Dimensional Actions with Statistically Dependent Components}
\label{hidim}

As mentioned in Sec.~\ref{probdet},
\xxx is of particular interest for high-dimensional  actions,
because SL can easily deal with those, while traditional RL does not.

Let us first consider the case of multiple trials, where
$out(k) \in \mathbb{R}^o$ encodes a probability distribution over
high-dimensional actions, where the $i$-th action component $out_i'(k)$ is either 1 or 0,
such that there are at most $2^o$ possible actions. 

C can be trained by Algorithm B2  to emit $out(k)$, 
given C's input history.
This is straightforward under the assumption that the components of $out'(.)$
are statistically independent of each other, given C's input history.

In general, however, they are not. 
For example, a C controlling a robot with 5 fingers should often send similar, 
statistically redundant commands to each finger,
e.g., when closing its hand.

To deal with this, Algorithms B1 and  B2  can be modified in a straightforward way.
Any complex high-dimensional action at a given time step can be computed/selected incrementally, 
component by component, where each component's probability also depends 
 on components already selected earlier. 

More formally, in Algorithm B1 
we can decompose each time step $t$ into $o$  discrete {\em micro time steps} 
 $\hat{t}(1),\hat{t}(2),\ldots,\hat{t}(o)$ 
(see~\cite{Schmidhuber:90diffgenau}, Sec.~on {\em ``more network ticks than environmental ticks''}).
At $\hat{t}(1)$ we initialize real-valued variable $out_{0}'(t)=0$. 
During $\hat{t}(i),1\leq i \leq o$, 
C computes $out_i(t)$, the probability of  $out_i'(t)$ being 1,
given C's internal state (based on its previously observed history) and its current
 inputs $all(t)$, $horizon(t)$, $desire(t)$, $extra(t)$ and $out_{i-1}'(t)$ (observed through an additional special 
  {\em action input unit} of C).
Then  $out_i'(t)$ is sampled accordingly,
and for $i < o$ used as C's new  {\em special action input} at the next micro time step $\hat{t}(i+1)$. 


Training of C in Step 2 of Algorithm B2 has to be modified accordingly. 
There are obvious, similar modifications of Algorithms B1 and  B2 for Gaussian and other 
types of probability distributions.

\subsubsection{Computational Power of RNNs: Generalization \& Randomness vs. Determinism}
\label{generalize}

This is an important subsection. First recall that
Sec.~\ref{probdet} pointed out how an FNN-based C of Algorithms A1/A2 in general will 
learn probabilistic policies even in deterministic environments,
since at a given time $t$, 
C can perceive only  the recent $all(t)$ but not the entire history $trace(t)$, 
reflecting an inherent Markov assumption~\cite{stratonovich1960,Schmidhuber:91nips,Kaelbling:96,Sutton:98,wiering2012}.

{\bf If there is only one single lifelong trial, however,}
this argument does not hold for the RNN-based C of Algorithms B1/B2,
because at each time step,  an RNN could in principle uniquely represent
the entire history so far, for instance, by learning to simply count the time steps~\cite{Gers:2000b}.

This is conceptually very attractive.
We do not even have to make any probabilistic assumptions any more.
Instead, \xxx
simply learns to map histories and commands directly to 
high-dimensional  deterministic actions $out'(.):=out(.) \in \mathbb{R}^{o}$.
(This tends to be hard for traditional RL.)

Even in seemingly probabilistic environments (Sec.~\ref{probenv}),
an RNN C could learn deterministic policies,
taking into account the precise histories after which these policies worked in the past,
assuming that what seems random actually may have been computed by some 
deterministic 
(initially unknown) algorithm, e.g.,  a pseudorandom number generator~\cite{Zuse:69,Schmidhuber:97brauer96,Schmidhuber:00v2,Schmidhuber:02ijfcs,Schmidhuber:02colt}. 


To illustrate the conceptual advantages of single life settings, let us consider a simple task where an agent can pass an obstacle either to the left or to the right, using continuous actions in [0,1] defining angles of movement, e.g., 0.0 means go left, 0.5 go straight (and hit the obstacle), 1.0 go right. 

First consider an episodic setting and a sequence of trials where C is reset after each trial. Suppose actions 0.0 and 1.0 have led to high reward 10.0 equally often, and no other actions such as 0.3 have triggered high reward. Given reward input command 10.0, the agent's RNN C will learn an expected output of 0.5, which of course is useless as a real-valued action---instead one has to somehow interpret this as an action {\em probability} based on certain assumptions about an underlying distribution (Sec.~\ref{det}, \ref{probenv}, \ref{hidim}). Note, however,  that the typical popular Gaussian assumptions would not make sense here.

On the other hand,  in a single life with, say, 10 subsequent sub-trials, C can learn arbitrary history-dependent algorithmic conditions of actions, e.g.: in trials 3, 6, 9, action 0.0 was followed by high reward. In trials 4, 5, 7, action 1.0 was. Other actions 0.4, 0.3, 0.7, 0.7 in trials 1, 2, 8, 10 respectively, yielded low reward. By sub-trial 11, in response to reward command 10.0, C should correctly produce either action 0.0 or 1.0 but not their mean 0.5. 

 In additional sub-trials, C might even  discover 
 complex conditions such as: if the  trial number is divisible by 3, then choose action 0.0, else 1.0. In this sense, in  single life settings, life is getting conceptually simpler, not harder. Because the whole baggage associated with probabilistic thinking and
 {\em a priori {\em  assumptions about probability distributions and environmental resets 
(see Sec. \ref{probenv}) is getting irrelevant and  can be ignored.

On the other hand,
C's success in case of similar commands in similar situations at different time steps will now all depend on its generalization capability.
For example, from its historic data, it must learn in step 2 of Algorithm B2 when precise time stamps are important and when to ignore them. 

Sure, even in deterministic environments, C might find it useful to invent a variant of
probability theory to model its uncertainty, and to make seemingly ``random'' decisions with the help of a self-invented deterministic internal pseudorandom generator. However, no probabilistic assumptions (such as the above-mentioned overly restrictive Gaussian assumption) should be imposed onto C a priori.

To improve C's generalization capability, well-known regularizers~\cite[Sec.~5.6.3]{888} 
can be used during training in Step 2 of Algorithm B2. See also Sec.~\ref{reducing}.

{\bf 
\xxx for RNNs or other general purpose computers 
without any probabilistic assumptions (Sec.~\ref{probdet}, \ref{probenv}, \ref{hidim})
may be both the simplest and most powerful \xxx variant. }

\subsubsection{RNNs With Memories of Initial Commands}
\label{commandmemory}

There are variants of \xxx with an RNN-based C that 
accepts commands such as {\em ``get so much reward per time in this trial"}  only in the beginning of each trial, 
or only at certain selected time steps, such that  $desire(.)$ and $horizon(.)$ do not  have to be updated any longer at every time step, because the RNN can learn to internally memorize previous commands. However, then C must also somehow be able to observe at which time steps $t$ to ignore $desire(t)$ and $horizon(t)$. This can be achieved through a special marker input unit whose activation as part of $extra(t)$ is 1.0 
only if the present $desire(t)$ and $horizon(t)$ commands should be obeyed (otherwise this activation is 0.0). 
Thus C can know during the trial: The current goal is to match the last command (or command sequence)  identified by this marker input unit. 
This approach can be implemented through obvious modifications of Algorithms B1 and B2. 

\subsubsection{Combinations with Supervised Pre-Training and Other Techniques}
\label{combine}

It is trivial to combine \xxx and SL, since both share the same basic framework. 
In particular, C can be pre-trained by SL to imitate teacher-given trajectories. 
The corresponding traces
can simply be added to C's training set of Step 2 of Algorithm B2.

Similarly, traditional RL methods or AI planning methods can be used to create
additional behavioral traces for training C.

For example, we may use the company NNAISENSE's  winner of the  NIPS 2017 ``learning to run"  competition to generate 
several behavioral traces of a successful, quickly running, simulated 3-dimensional skeleton
controlled through relatively high-dimensional actions, 
in order to pre-train and  initialize C.
C may then use  \xxx  to further refine its behavior. 

\section{Compress Successful Behaviors Into a Compact Standard Policy Network Without Command Inputs}
\label{compress}

C has to learn a possibly complex mapping from desired rewards, time horizons, and normal sensory inputs, 
to actions. Small changes in initial conditions or reward commands may require quite different actions.  
A deep and complex network may be
necessary to learn this. During exploitation, however, we do not need this complex mapping any longer, we just need
a working policy that maps sensory inputs to actions. This policy may fit into a much smaller network. 

Hence, to exploit successful behaviors learned through algorithms A1/A2 or B1/B2, 
we simply compress them into a policy network called CC,
like in the 1991 chunker-automatizer system~\cite{chunker91and92}, 
where a student net (the ``automatizer'') is continually re-trained 
not only on its previous skills (to avoid forgetting), but also
 to imitate the behavior of a teacher net  (the ``chunker''), 
which itself keeps learning new behaviors.
The {\sc PowerPlay}  framework~\cite{powerplay2011and13,Srivastava2013first} also
uses a similar approach, learning one  task after another, using  environment-independent replay of behavioral traces (or functionally equivalent but more efficient approaches) to avoid forgetting previous skills
and to compress or speed up
previously found, sub-optimal solutions,
e.g.,~\cite[Sec.~3.1.2]{powerplay2011and13}.
Similar for the ``One Big Net''~\cite{onebignet2018} and a
recent study of incremental skill learning with feedforward networks~\cite{progressive2018}.

Using the notation  of Sec.~\ref{formal},
the policy net CC is like C, 
but without special input units for 
the command inputs $horizon(.)$, $desire(.)$, $extra(.)$.
We immediately consider the case where CC is an RNN living in a partially observable environment (Sec.~\ref{partial}).

\subsection*{Algorithm Compress (replay-based training on previous successful behaviors):}
\label{algcompress}

\begin{enumerate}

\item  
{\bf For} each previous trial that is considered successful: Using the notation  of Sec.~\ref{formal}, 
{\bf For} $1 \leq k \leq T$ {\bf do}: 
Train RNN CC
to emit action $out'(k)$ at time $k$  in response to the previously observed part of the history $trace(k-1)$.

\end{enumerate}

For example, in a given environment, \xxx can be used to solve an RL task requiring to achieve maximal reward / minimal time
under particular initial conditions (e.g., starting from a particular initial state). 
Later, Algorithm Compress can collapse many different satisfactory solutions for many different initial conditions
 into CC, which ignores reward and time commands.

\newpage
\section{Imitate a Robot, to Make it Learn to Imitate You!}
\label{imitate}

The concept of learning to use rewards and other goals as command inputs has broad applicability.
In particular, we can apply it in an elegant and straighforward way to train robots 
on {\em learning by demonstration} tasks~\cite{visualdemo1997,schaal1999,argall2009survey,duan2017,levine2017} considered notoriously difficult in traditional robotics.  
We'll conceptually simplify an approach~\cite{levine2017} 
for teaching
a robot to imitate humans.

For example,
suppose that an RNN C should learn to control a complex 
humanoid robot with eye-like cameras perceiving a
visual input stream.
We want to teach it complex tasks, 
such as assembling a smartphone, 
solely by visual demonstration,
without touching the robot - a bit like we'd teach a kid.

First the robot must learn what it means to imitate a human. 
Its joints and hands may be quite different from yours. 
But you can simply let the robot execute already known or even accidental behavior.
Then simply imitate it with your own body! 
The robot tapes a video of your imitation  through its cameras.
The video is used as a sequential command input  for the RNN controller C 
(e.g., through parts of $extra()$, $desire()$, $horizon()$),
and C is trained by SL to respond with its known, already executed behavior. 
That is, C can learn by SL to imitate you, because you imitated C.

Once C has learned to imitate or obey several video commands like this,
let it generalize: do something it has never done before,
and use the resulting video as a command input.

In case of unsatisfactory imitation behavior by C,
imitate it again, to obtain additional training data.
And so on, until performance is sufficiently good. 
The algorithmic framework {\em  Imitate-Imitator}
formalizes this procedure.

\subsection*{Algorithmic Framework: Imitate-Imitator}
\label{algimit}
\begin{enumerate}

\item {\bf Initialization:} 
Set temporary integer variable $i:=0$. 

\item {\bf Demonstration:} 
Visually show to the robot what you want it to do, 
while it videotapes your behavior,
yielding a video $V$. 

\item {\bf Exploitation / Exploration:} 
Set $i:=i+1$.
Let RNN C sequentially observe $V$ and then produce a trace $H^i$ of 
a series of interactions with the environment
(if in exploration mode, produce occasional random actions).
If the robot is deemed a satisfactory imitator of your behavior, exit.

\item {\bf Imitate Robot:} 
Imitate $H^i$  with your own body, while
the robot records a video $V^i$ of your imitation.

\item {\bf Train Robot:} 
For all $k, 1 \leq k \leq i$ train RNN C through 
gradient descent~\cite[Sec.~5.5]{888}
 to sequentially observe $V^k$ (plus the already known total vector-valued cost $R^k$ of $H^k$)
 and then produce $H^k$, where the pair
$(V^k,R^k)$ is interpreted as a sequential command to perform $H^k$ under cost $R^k$.
Go to Step 3 (or to Step 2 if you want to demonstrate anew).

\end{enumerate}

It is obvious how to implement variants of this procedure 
through straightforward modifications of Algorithms B1 and B2
along the lines of 
Sec.~\ref{predicates}, e.g., using a gradient-based sequence-to-sequence mapping
approach based on LSTM, e.g.,~\cite{Graves:09tpami,sutskever2014,wu2016google}.

Of course, the {\em  Imitate-Imitator} approach is not limited to videos.
All kinds of sequential, possibly multi-modal sensory data could be used 
to describe desired behavior to an RNN C,
including spoken commands, or gestures.
For example, observe a robot, then describe its behaviors in your own language,
through speech or text.
Then let it learn to map your descriptions to its own corresponding behaviors.
Then describe a new desired behavior to be performed by the robot,
and let it generalize from what it has learned so far.

Once the robot has learned to execute command
$(V^k,R^k)$ through  behavior $H^k$,  standard \xxx without a teacher 
can be used to further refine $H^k$, by commanding the robot to produce similar behavior 
under different cost $\hat{R}^k$ (of the same dimensionality as $R^k$).
If necessary, the robot is trained to obey the commands through an additional series of trials. 
For example, a robot that already knows how to assemble some object
may now learn by itself to assemble it faster or with less energy. 

The central  idea of the present Sec.~\ref{imitate} on what we'd like to call {\em show-and-tell robotics} or
{\em watch-and-learn robotics} or  {\em see-and-do robotics} 
may actually explain why biological evolution has evolved parents who imitate the babbling of their babies:
the latter can thus quickly learn to translate input sequences caused by the behavior of their parents into 
action sequences corresponding to their own equivalent behavior.
Essentially they are learning their parent's language to describe behaviors, 
then generalize and translate 
previously unknown behaviors of their parents into equivalent own behaviors.

\section{Relation of Upside Down RL  to Previous Work}
\label{previous}

Using SL for certain aspects of RL dates back to the 
1980s and 90s~\cite{Werbos:87,Munro:87,Jordan:88,Werbos:89identification,Werbos:89neurocontrol,RobinsonFallside:89,NguyenWidrow:89}.
In particular, like  \xxxn, our early  end-to-end-differentiable recurrent RL machines (1990) also observe 
vector-valued reward signals  as sensory inputs~\cite{Schmidhuber:90diffgenau,Schmidhuber:90sandiego,Schmidhuber:91nips}.
What is the concrete difference between those and  \xxxn?
The earlier systems~\cite{Schmidhuber:90diffgenau,Schmidhuber:90sandiego,Schmidhuber:91nips} also use gradient-based SL in RNNs
 to learn mappings from costs/rewards 
and other inputs to actions. But unlike \xxx they do not have {\em desired} rewards as {\em command} inputs, and typically the training 
depends on an RNN-based predictive world model M (which predicts rewards, among other things)
to compute gradients for the RNN controller C. 
\xxxn, however, does not depend at all on good reward predictions (compare~\cite[Sec.~5]{learningtothink2015}),
only on the generalization ability of the learned mapping from  previously observed rewards and other inputs to action probabilities. 

What is the difference to  our early multi-goal RL systems (1990) which also had extra input vectors
used to encode possible goals~\cite{SchmidhuberHuber:91}?
Again, it is essentially the one mentioned in the previous paragraph: \xxx does not require
additional  predictions of reward.

What is the difference to  our early end-to-end-differentiable  
hierarchical RL (HRL) systems (1990)  which also had extra task-defining inputs in form of start/goal combinations,
learning to invent sequences of  subgoals~\cite{Schmidhuber:91icannsubgoals}? 
Unlike \xxxn,
such HRL also needs a predictor of costs/rewards (called an evaluator), given start/goal combinations,
to derive useful subgoals through gradient descent.

What is the difference to  hindsight experience replay (HER, 2017)~\cite{hindsight2017} 
extending experience replay (ER, 1991)~\cite{Lin:91}?
HER replays paths to randomly encountered potential goal locations,
but still depends on traditional RL.
HER's controller neither sees extra  real-valued $horizon$ and $cost$ inputs 
nor general computable predicates thereof,
and thus 
does not learn to generalize from {\em known} costs in the training set 
to {\em desirable} costs in the generalization phase.
(HER also does not use an RNN to deal with partial observability through encoding 
the entire history). Similar considerations hold for hindsight policy gradients~\cite{rauber2017hindsight}.

What is the difference to RUDDER~\cite{rudder2019} which also uses gradient-based SL in RNNs to perform contribution analysis, mapping rewards to state-action pairs? Unlike \xxxn,
RUDDER does not use desired rewards as command input for an SL model.

To summarise, as discussed above, 
mapping rewards~\cite{Schmidhuber:90diffgenau,Schmidhuber:90sandiego,Schmidhuber:91nips} 
and goals~\cite{SchmidhuberHuber:91}
 (plus other inputs) to actions is not new.
But traditional RL methods~\cite{Kaelbling:96,Sutton:98,wiering2012} 
do not have  {\em command} inputs in form of {\em desired} rewards, and most of them 
 need some additional method for learning to select actions 
based on predictions of future rewards. For example, 
a more recent system~\cite{dosovitskiy2017} also predicts future measurements (possibly rewards), given actions, 
and selects actions leading to best predicted measurements, given goals. 
 A characteristic property  of \xxxn, however, is its very simple shortcut: 
 it learns directly from (possibly accidental) experience the mapping from rewards to actions.

\xxx  is also very different from traditional 
black box optimization (BBO)~\cite{Rechenberg:71,Schwefel:74,Holland:75,Fogel:66} such as neuroevolution~\cite{miller:icga89,Sims:1994:EVC,gomez:phd,glasmachers:2010b}
which  can be used to solve complex RL problems
in partially observable environments~\cite{Gomez:08jmlr}
through iterative discovery of better and better parameters  of an adaptive controller,
yielding more and more reward per trial. 
\xxx does not even try to modify 
any weights with the objective of  increasing reward.
Instead it just tries to understand from previous experience through standard gradient-based learning 
how to translate (desired) rewards etc into corresponding  actions.
Unlike BBO, 
\xxx  is also applicable when there is only one single lifelong trial;
the new observations of any given time step can immediately be used 
to improve the learner's overall behavior.

What is the difference between \xxx and   {\sc PowerPlay}  
(2011)~\cite{powerplay2011and13,Srivastava2013first}? Like \xxxn,
 {\sc PowerPlay}  does receive extra   command inputs
in form of arbitrary (user-defined or self-invented) computable task specifications,
 possibly 
involving start states, goal states, and costs including time. It even orders (at least the self-invented)
  tasks automatically by the 
 computational difficulty of adding their solutions to the skill repertoire.
 But it does not necessarily systematically 
consider all previous training sequences between all 
possible pairs  of previous time steps encountered so far by accident. 
See also Sec.~\ref{infinite}.

Of course, we could limit  {\sc PowerPlay}'s choice of new problems to problems of the form:
{\em choose a unique new command for C reflecting a
computable predicate that is true for some already observed action sequence (Sec.~\ref{infinite}),
and add the corresponding skill to C's repertoire, without destroying previous knowledge.}
Such an association of a new command with a corresponding skill or policy will cost time and other resources;
{\sc PowerPlay} will, as always, prefer new skills that are easy to add. 
(Recall that one can train C only on finitely many commands, which should be chosen wisely.)

Note also that at least the strict versions of  {\sc PowerPlay}  insist that adding a new skill does not decrease performance
on (replays of) previous tasks, while \xxxn's occasional sychronization of Algorithms  
A1/A2 and B1/B2 
does not immediately guarantee this, due to limited time between synchronizations, 
and basic limitations of gradient descent. 
Nevertheless, in the long run, 
Algorithms A2/B2 of \xxx will keep up with the stream of  incoming new observations from Algorithms A1/B1, 
and thus won't forget previous skills of C due to constant retraining, much like {\sc PowerPlay}.

\section{Experiments}
\label{exp}

A separate paper~\cite{srivastava2019}  describes the
concrete implementations used in our first experiments with 
 a pilot version of \xxxn, and presents remarkable  experimental results.

\section{Conclusion}
\label{conclusion}

Traditional RL predicts rewards, and uses a myriad of methods for translating
those predictions into good actions. 
\xxx shortcuts this process, creating a direct 
mapping from rewards, time horizons and other inputs to actions.
Without depending on reward predictions,
and without explicitly maximizing expected rewards,
 \xxx simply learns by gradient descent to map task specifications or commands 
 (such as: {\em get lots of reward within little time}) to action probabilities.
 Its success depends on the generalization abilities of deep / recurrent neural nets.
 Its potential drawbacks are essentially those of traditional gradient-based learning: local minima, 
 underfitting, overfitting, etc.~\cite{bishop:2006,888}.
 Nevertheless, experiments in a separate paper~\cite{srivastava2019}  show that even our initial pilot version of  \xxx  can outperform
traditional RL methods on certain challenging problems.

A  related {\em  Imitate-Imitator} approach is to 
imitate a robot, then let it learn to map its observations of the imitated behavior to its own behavior,
then let it generalize, by demonstrating something new, to be imitated by the robot.

\section{Acknowledgments} 
\label{ack}

I am grateful to 
Paulo Rauber, 
Sjoerd van Steenkiste, 
Wojciech Jaskowski, 
Rupesh Kumar Srivastava,
Jan Koutnik, 
Filipe Mutz,
and Pranav Shyam  
for useful comments. 
This work was supported in part by 
a European Research Council Advanced Grant (no: 742870).

\bibliography{bib}
\bibliographystyle{abbrv}
\end{document}